\documentclass[journal]{IEEEtran}

\usepackage{graphicx}
\usepackage{amsmath}
\usepackage{amssymb}
\usepackage{booktabs}
\usepackage{subcaption}

\usepackage{multirow}
\usepackage{xcolor}

\ifCLASSINFOpdf
\else
\fi
\begin{document}
%
\title{Enhancing Power Grid Inspections with Machine Learning}
%
%
%

\author{Diogo~Lavado, 
        Ricardo~Santos,
        André~Coelho,
        João~Santos,
        Alessandra~Micheletti,
        and~Cláudia~Soares,
\thanks{D. Lavado is with the Department of Computer Science of NOVA University of Lisbon, Portugal, along with the Department of Environmental Sciences of the University of Milan, Italy e-mail: d.lavado@campus.fct.unl.pt}
\thanks{R. Santos and A. Coelho are with Labelec, Lisbon, Portugal}
\thanks{J. Santos is with the NEW Center of Research, Lisbon, Portugal}
\thanks{A. Micheletti is with the Department of Environmental Sciences of the University of Milan, Italy}
\thanks{C. Soares is with the Department of Computer Science of NOVA University of Lisbon, Portugal}
\thanks{Manuscript received April 19, 2005; revised August 26, 2015.}}

\maketitle

\begin{abstract}
Ensuring the safety and reliability of power grids is critical as global energy demands continue to rise. Traditional inspection methods, such as manual observations or helicopter surveys, are resource-intensive and lack scalability. This paper explores the use of 3D computer vision to automate power grid inspections, utilizing the TS40K dataset—a high-density, annotated collection of 3D LiDAR point clouds. By concentrating on 3D semantic segmentation, our approach addresses challenges like class imbalance and noisy data to enhance the detection of critical grid components such as power lines and towers. The benchmark results indicate significant performance improvements, with IoU scores reaching 95.53\% for the detection of power lines using transformer-based models. Our findings illustrate the potential for integrating ML into grid maintenance workflows, increasing efficiency and enabling proactive risk management strategies.
\end{abstract}


%
\IEEEpeerreviewmaketitle

\section{Introduction}
\IEEEPARstart{E}{nsuring} the safe and efficient operation of electrical transmission and distribution systems is a fundamental responsibility for power grid operators. Regular inspections are vital to maintaining grid reliability by detecting structural defects, assessing collision risks, and preventing environmental hazards such as vegetation encroachment~\cite{ahmad2013vegetation}, damages to the structure, and severe weather events such as wildfires~\cite{zamuda2013us,folga2016national,muhs2019stochastic}. 
However, traditional inspection methods that rely on on-site personnel or manned helicopters are resource-intensive, expensive, and time-consuming. As transmission networks keep expanding in size and complexity, these methods find it difficult to deliver the efficiency and scalability needed to tackle emerging challenges.
Recently, the use of unmanned aerial vehicles (UAVs) equipped with LiDAR sensors has transformed the way power grid inspections are conducted~\cite{toth2010field,jenssen2019intelligent,arreola2018improvement}. UAVs can remotely capture high-resolution 3D point cloud representations of transmission and distribution systems, enabling operators to inspect the grids without deploying personnel on site~\cite{toth2010smart}. 
However, significant inefficiencies remain. Maintenance teams must still manually annotate large datasets and evaluate risks, a process that is tedious and susceptible to overlooking critical issues that require immediate attention.
Using 3D data, 3D Computer Vision (CV) presents itself as a viable option to automate and improve power grid inspections.
Specifically, 3D semantic segmentation allows for the automatic identification of key scene elements, such as power lines, towers, and vegetation, directly from LiDAR point clouds. 
This eliminates the need for manual annotations while supporting proactive maintenance strategies to address risks such as equipment failures or environmental hazards. This significantly boosts inspection efficiency, minimizes costs, and guarantees timely interventions, paving the way for safer and more reliable grid operations.

While 2D imagery has been widely used for power grid inspections~\cite{jalil2019fault,hosseini2020intelligent,maduako2022deep,santos2024uav,chandaliya2024uav,chatzargyros2024uav}---leveraging RGB or thermal images for fault detection, component identification, and vegetation encroachment---it comes with significant limitations. These methods rely on favorable weather and lighting conditions, making them unreliable in real-world scenarios where shadows, glare, mist or low visibility can obscure critical grid elements. Moreover, the lack of depth information hinders accurate distance estimation between power lines and vegetation, reducing their effectiveness in safety assessments.
3D data overcomes these challenges by providing geometric and spatial context, enabling more precise analysis. To this end, the TS40K~\cite{lavado2024ts40k} dataset is a strong candidate for training machine learning models tailored to power grid inspections. It offers high-density UAV-captured 3D point clouds of rural environments, featuring diverse medium- and small-voltage towers, detailed annotations for power lines, towers, and vegetation, and consideration of irregular terrains. These attributes make TS40K particularly well-suited for training machine learning models to meet the demands of automated, comprehensive, and reliable power grid maintenance.


Building on these challenges, our research tackles the significant gap in automating power grid inspections through the use of 3D data and machine learning techniques. The power of transformer-based models, trained on the TS40K dataset facilitates a thorough understanding of these systems. This paper presents a novel inspection pipeline that capitalizes on the unique attributes of TS40K to train state-of-the-art 3D semantic segmentation models, concentrating on essential inspection tasks such as vegetation encroachment detection and infrastructure evaluation. Our contributions are as follows:
\begin{enumerate}
    \item An in-depth analysis of 3D semantic segmentation methods used for power grid inspections, emphasizing their strengths and limitations.
    \item A benchmarking study on TS40K that incorporates innovative features to enhance model performance.
    \item A proposed inspection tool that combines ML predictions with manual review processes to ensure reliability.
\end{enumerate}

The remainder of this paper is organized as: Section~\ref{sec:related_work} discusses related work in power grid inspections and 3D semantic segmentation. Section~\ref{sec:tasks_benchmarks} presents the tasks and benchmarks defined for the TS40K dataset. Section~\ref{sec:inspection_tool} introduces our proposed inspection tool. Finally, Section~\ref{sec:conclusions} concludes with a discussion on implications and future work.

\section{Related Work~\label{sec:related_work}}

\subsection{Power Grid Inspection Methodologies}

Traditional power grid inspections are primarily conducted manually, depending on on-site maintenance personnel or manned helicopters to visually inspect transmission lines and supporting infrastructure~\cite{lavado2024ts40k}. 
While effective, these methods are resource-intensive, expensive, and often susceptible to human error, particularly when inspecting large and intricate terrains. As transmission grids keep expanding, there is an urgent need for more efficient and scalable inspection methodologies.
3D semantic segmentation offers an efficient and reliable solution. By leveraging 3D point clouds captured by unmanned aerial vehicles (UAVs) equipped with LiDAR sensors, CV models can automate the identification of 3D points belonging to critical elements such as power lines, support towers, and surrounding vegetation. 
This segmentation allows for the detection of structural failures, potential vegetation intrusions, and areas requiring immediate inspection. This greatly reduces the dependence on manual labor. 
Automating these processes accelerates inspection workflows and mitigates risks associated with delayed maintenance, such as power outages or wildfire hazards.


\subsection{3D semantic Segmentation}

3D semantic segmentation involves dividing a 3D point cloud into distinct regions labeled with meaningful categories. This task is particularly relevant for power grid inspections, where identifying components such as power lines, support towers, and vegetation can greatly enhance maintenance efficiency and reliability. Research in this field can be categorized into four paradigms: \textit{projection-based}, \textit{discretization-based}, \textit{point-based}, and \textit{hybrid methods}.
Projection-based approaches~\cite{su2015multi,lawin2017deep,yang2019learning,lyu2020learning} convert 3D point clouds into 2D representations, utilizing the strengths of established convolutional neural networks (CNNs). However, this process risks losing vital geometric details, such as depth.
Discretization-based methods~\cite{choy20194d,zhou2018voxelnet,le2018pointgrid,meng2019vv,zhang2020polarnet} process 3D data by dividing the space into a grid of small cubes, or voxels, which maintains certain spatial structure but is computationally expensive for high-resolution data.
In contrast, point-based techniques~\cite{qi2017pointnet,qi2017pointnet++,li2018pointcnn,thomas2019kpconv,hu2020randla,kong2023rethinking,lai2023spherical,zhao2021point,wu2022point,wu2023ptv3,pointcept2023} directly operate on raw 3D points, preserving fine-grained geometric information and achieving state-of-the-art performance on various benchmarks.
Transformer-based models have recently advanced point-based segmentation by leveraging self-attention mechanisms to capture both local and global dependencies within 3D data. Architectures such as Point Transformer~\cite{zhao2021point}, Point Transformer V2~\cite{wu2022point}, and Point Transformer V3~\cite{wu2023ptv3} improve feature aggregation by dynamically weighting point relationships, allowing for better geometric reasoning. These models consistently achieve state-of-the-art results on 3D benchmarks, including TS40K, demonstrating their effectiveness in power grid inspection tasks where fine structural details are critical.
Hybrid methods~\cite{dai20183dmv,jaritz2019multi,tang2020searching,hou2022point,liu2023uniseg} combine various approaches to capture complementary features, enhancing overall performance.

\subsection{TS40K Dataset}
\begin{figure*}[t]
\centering
\begin{minipage}[t]{0.48\textwidth} 
    \begin{subfigure}[b]{\textwidth}
        \includegraphics[width=\linewidth]{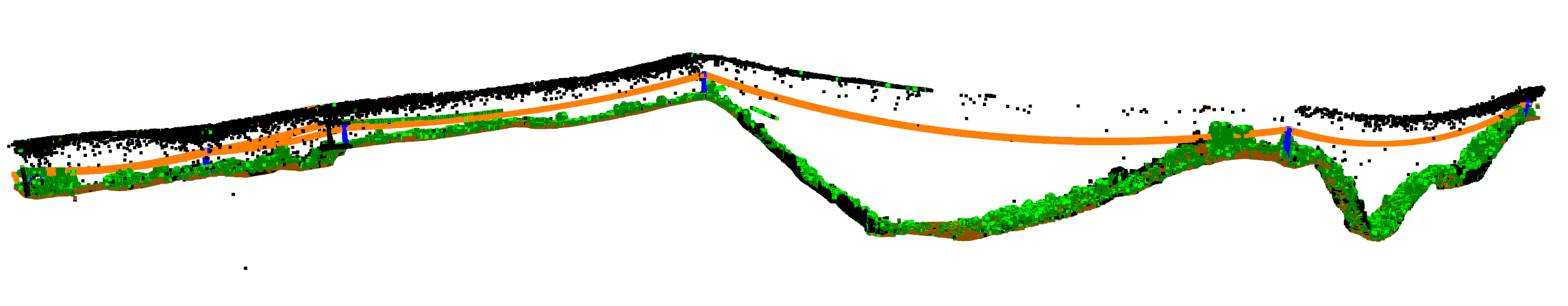}
        \caption{Raw TS40K sample}
        \label{fig:teaser-raw-sample}
    \end{subfigure}
    \vfill
    \begin{subfigure}[b]{0.33\textwidth}
        \includegraphics[width=\linewidth]{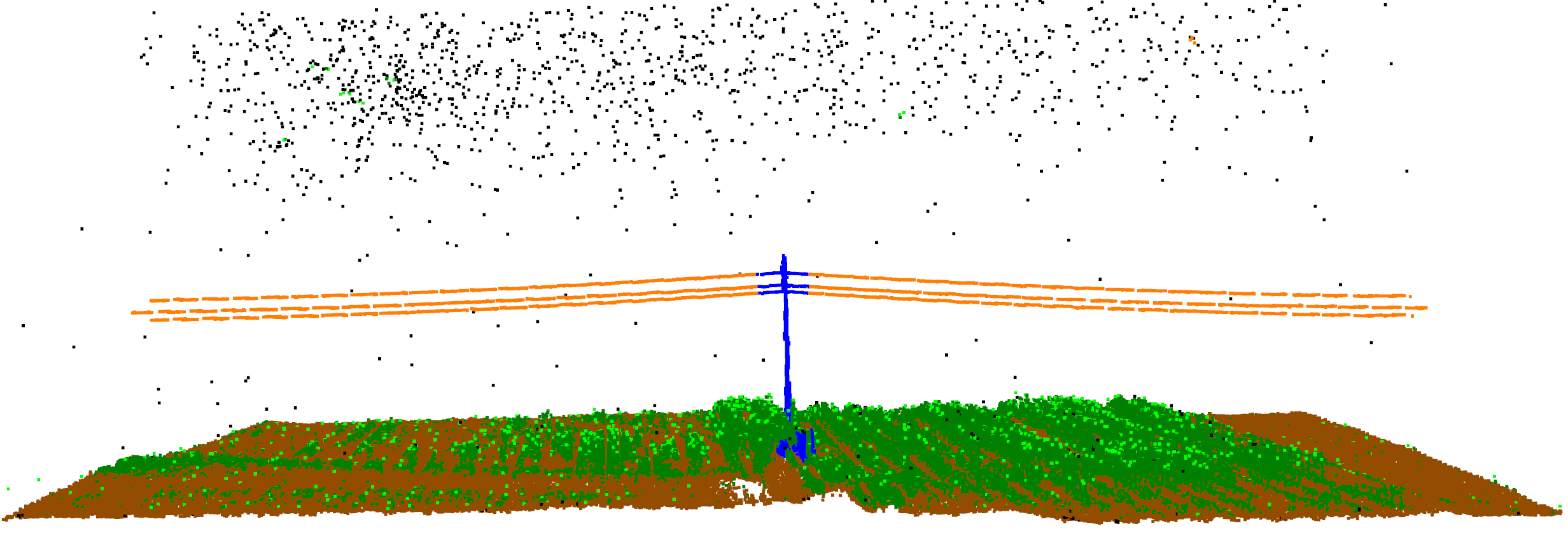}
        \caption{Tower-radius}
        \label{fig:teaser-tower-radius}
    \end{subfigure}
    \hfill
    \begin{subfigure}[b]{0.33\textwidth}
        \includegraphics[width=\linewidth]{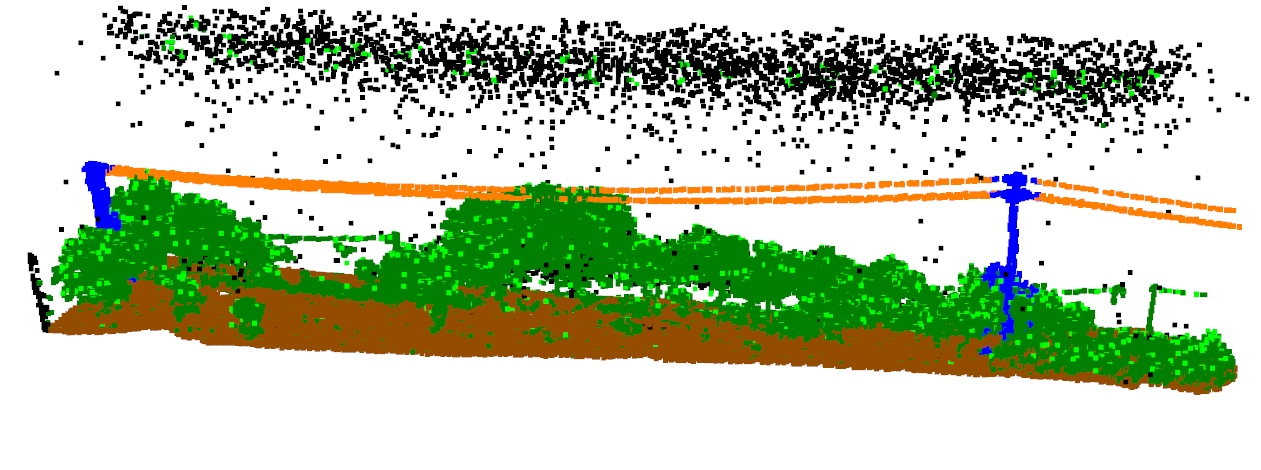}
        \caption{Power-line}
        \label{fig:teaser-power-line}
    \end{subfigure}
    \hfill
    \begin{subfigure}[b]{0.30\textwidth}
        \includegraphics[width=\linewidth]{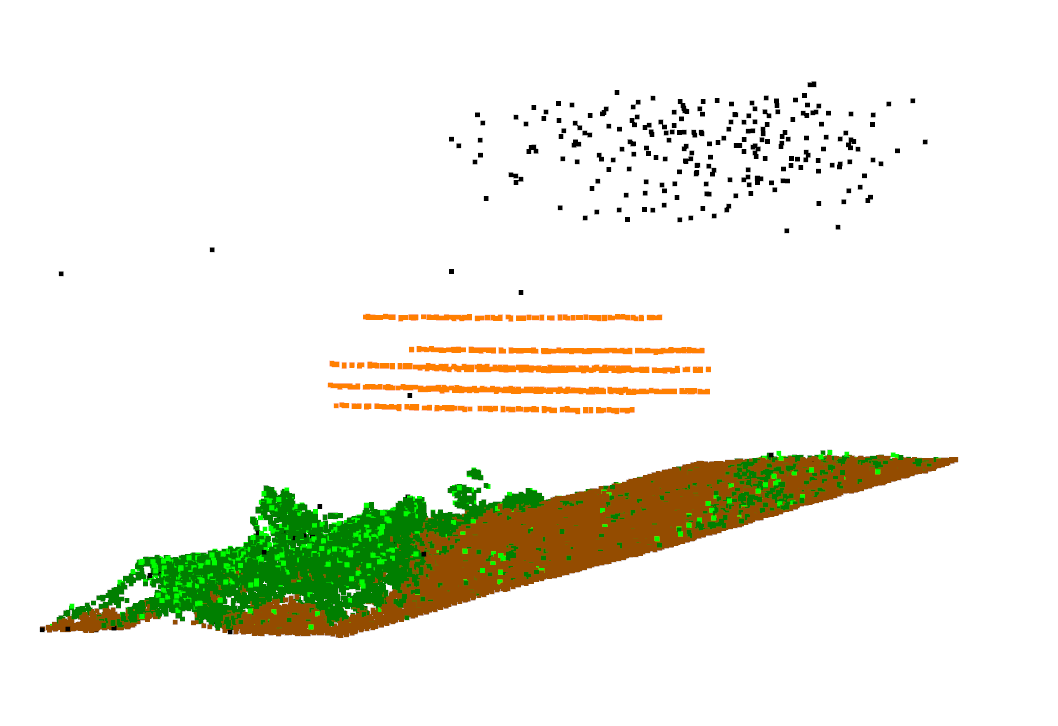}
        \caption{No-tower}
        \label{fig:teaser-no-ts}
    \end{subfigure}
    \vfill
    \includegraphics[width=\linewidth]{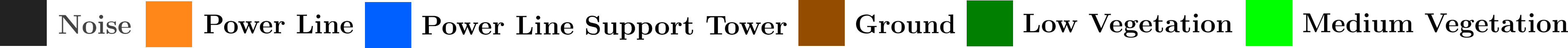}
\end{minipage}
\quad
\begin{minipage}[t]{0.48\textwidth} 
    \begin{subfigure}[b]{1.0\textwidth}
        \includegraphics[width=\linewidth]{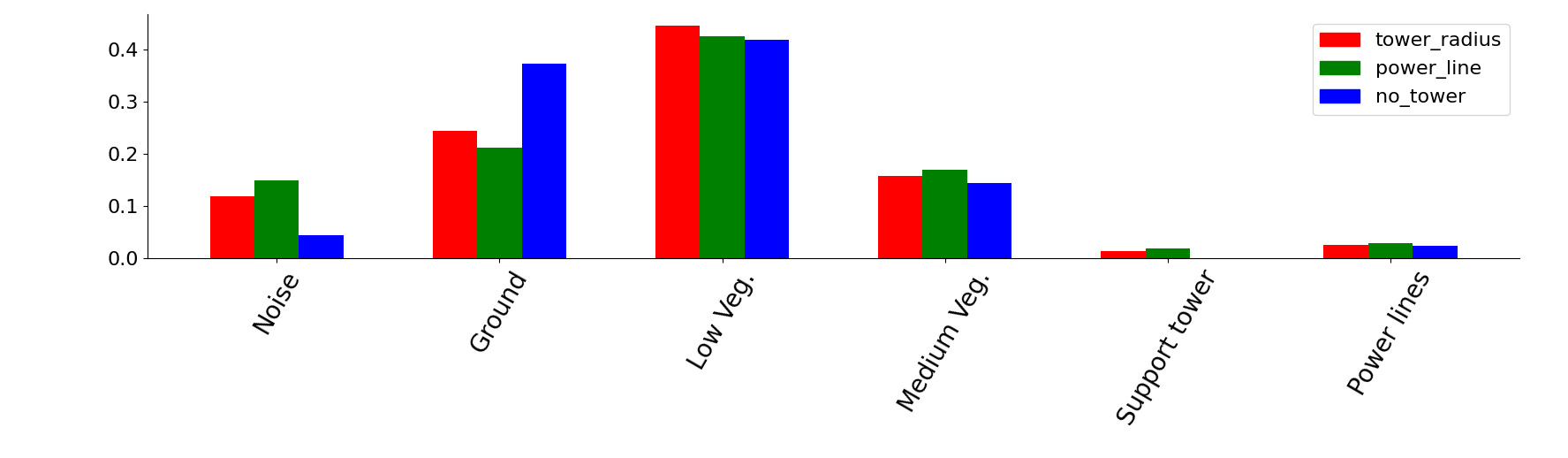}
        \caption{Sample type density}
        \label{fig:density1}
    \end{subfigure}
    \hfill
    \begin{subfigure}[b]{1.0\textwidth}
        \includegraphics[width=\linewidth]{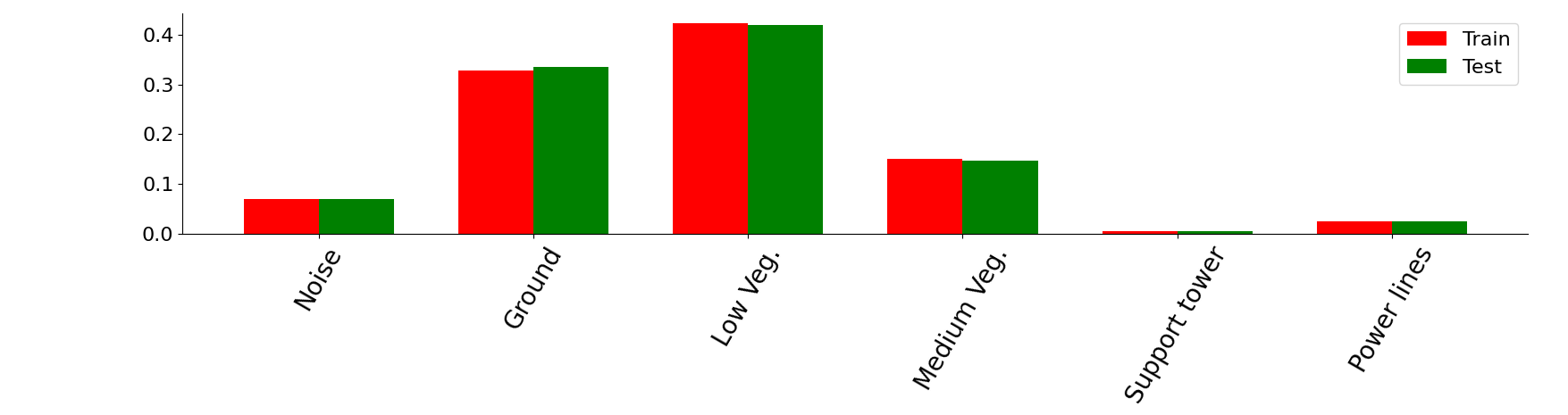}
        \caption{Overall train/test class density}
        \label{fig:density2}
    \end{subfigure}
\end{minipage}

\caption{The TS40K dataset is derived from raw 3D scans illustrated in Figure~\ref{fig:teaser-raw-sample} and processed into three different sample types: \textit{(1) Tower-radius} focuses on the towers that support power lines and its environment (Fig.~\ref{fig:teaser-tower-radius}). \textit{(2) Power-line} samples have power lines as their main focus in the 3D scenes (Fig.~\ref{fig:teaser-power-line}). \textit{(3) No-tower} samples represent rural terrain where the transmission system is located, excluding supporting towers but potentially including power lines (Fig.~\ref{fig:teaser-no-ts}).
In Figures~\ref{fig:density1} and~\ref{fig:density2}, we showcase the semantic class densities of the TS40K dataset. Figure~\ref{fig:density1} illustrates the class density for each of the sample types and Figure~\ref{fig:density2} shows the overall class density in the TS40K train and test sets.}
\label{fig:teaser}
\end{figure*}

The TS40K dataset provides a novel resource for advancing 3D scene understanding specifically related to electrical transmission systems, a domain that has been mostly neglected by existing datasets. 
TS40K is captured using drone-mounted LiDAR sensors and comprises high-density point clouds covering over 40,000 km of electrical transmission systems. Each scan is geo-referenced and annotated with 1 out of 5 inspection labels. 
The dataset offers unique characteristics, including uniform density, high spatial precision (2.5 cm), and a lack of occlusion, making it particularly suitable for the detailed analysis required in maintenance applications.
The 22 annotated labels are intended to streamline inspections rather than model training, so they are grouped to five semantically meaningful classes. These classes represent critical scene elements necessary for evaluating the risk of contact between the electrical system and its surroundings, specifically ground, low and medium vegetation, power line support towers, and power lines.
Additionally, the dataset presents three distinct sample types: tower-radius, which focuses on tower environments; power-line, which emphasizes power lines between towers; and no-tower, which captures terrain without visible transmission structures. This categorization enables tailored model evaluations based on specific inspection scenarios.

TS40K addresses the challenges of power grid inspection by replicating real-world complexities, such as class imbalance, sensor noise, and mislabeled annotations. For example, while ground and vegetation account for over 90\% of the dataset, power line support towers and power lines make up less than 2\%, presenting significant difficulties for machine learning algorithms. 
Additionally, the drone-collected data inherently includes spurious noise, particularly under adverse environmental conditions, which can obscure critical structures. 
The dataset also incorporates annotation noise, reflecting practical constraints faced by human inspectors during labeling.
Benchmarking on TS40K highlights significant performance gaps in current state-of-the-art segmentation models. 
Despite achieving strong results in urban datasets, methods like KPConv~\cite{thomas2019kpconv} and Point Transformer V2~\cite{wu2022point} exhibit low performance in power grid-related tasks, particularly in detecting power line support towers with 43\% Intersection over Union (IoU).
The authors attribute this performance lag to the diverse structures that compose supporting towers along with their low point-density in the dataset. Power lines on the other hand, enjoy a 94\% IoU despite having similar point-densities.
These results highlight the need for further research to develop robust, noise-resilient models tailored to this application domain.

\subsubsection{3D Datasets on Transmission Networks}
Several datasets exist that capture aspects of rural or forest terrain, as well as power grid elements, but each has specific limitations when applied to power grid inspections. Forest3D~\cite{trochta20173d} is centered solely on the 3D representation of tree crowns, emphasizing their instance segmentation without consideration for power grid components. 
GTASynth~\cite{curnis2022gtasynth}, a synthetic dataset of non-urban environments, features low-density point clouds with significant object occlusion, captured from a vehicle’s perspective that differ significantly from UAV-based grid inspections. 
Similarly, NEON~\cite{marconi2019data} focuses on airborne LiDAR data to predict tree crown dimensions but does not annotate grid infrastructure. DALES~\cite{varney2020dales}, while including high-voltage towers, is limited to urban environments and lacks the diversity found in rural grids.
In contrast, TS40K~\cite{lavado2024ts40k} specifically targets electrical power grids in rural areas, including medium and small-voltage towers that are more varied in shape and harder to distinguish from their surroundings.

\subsection{Power Line Detection Methods}

Power grid inspection has traditionally relied on on-site maintenance personnel and manned helicopters, where the grid is examined visually or with portable devices. These methods are labor-intensive, costly, and prone to inefficiencies, emphasizing the need for process automation. To address this, UAVs equipped with LiDAR sensors are increasingly being deployed to capture detailed 3D point cloud representations of power grid environments.

In the work of Ding et al.~\cite{ding2021electric}, simultaneous localization and mapping (SLAM) algorithms are combined with multi-sensor data to enable UAV-based patrols of electrical grids. This approach employs a multi-view-based technique for point cloud segmentation, but the reliance on 3D reconstructions derived from 2D raster maps often leads to significant information loss and reduced accuracy. 
Similarly, Guo et al.~\cite{guo2019research} project point clouds onto the $xy$-plane to cluster and segment power lines. However, this method overlooks critical contextual information, such as ground features and irregular terrain, and focuses primarily on incomplete segments of power lines. Alternative approaches, such as those presented by Tao et al.~\cite{tao2019study}, leverage fine-grained elevation statistics of the original point cloud in combination with $xy$-plane projections to enhance segmentation accuracy.

While these methods advance the automation of power line detection, they predominantly target high-voltage power lines and fail to account for the supporting towers, which are essential for comprehensive inspections. 
They also neglect other critical scene elements, such as vegetation, which poses a significant risk of contact with the grid and must be assessed to ensure operational safety.
Another major limitation of these approaches is the lack of publicly available datasets tailored for comprehensive power line inspections. Many methods rely on proprietary or restricted datasets, hindering reproducibility and further development.

\section{Tasks and Benchmarks~\label{sec:tasks_benchmarks}}

\subsection{3D Semantic Segmentation vs. 3D Object Detection in Power Grid Inspection}
3D semantic segmentation operates at the point level, assigning a semantic class to every point in the LiDAR scan. This level of granularity is particularly useful for identifying critical infrastructure components, such as power lines, support towers, and surrounding vegetation, and for assessing risks like vegetation intrusion or structure degradation. 
In contrast, 3D object detection simplifies the task by predicting bounding boxes for objects of interest. While this approach is computationally less intensive and can be effective for identifying well-defined objects, it lacks the precision needed to evaluate finer details, such as the exact location of a fault on a power line or the extent of vegetation growth near the grid.
For power grid inspections, where safety and reliability are paramount, the detailed insights provided by 3D semantic segmentation often outweigh the simplicity of object detection.

\subsection{Metrics}

In the evaluation of 3D semantic segmentation models, the confusion matrix is a fundamental tool that organizes the outcomes of a model's predictions into distinct categories:
\begin{itemize}
    \item \textbf{True Positives (TP)}: represent points correctly predicted as belonging to a specific class. High TP rates indicate that the model effectively identifies critical elements.
    \item \textbf{False Positives (FP)}: occur when the model incorrectly assigns points to a class they do not belong to. In the context of power grid inspections, FP errors could mean mistakenly labeling vegetation as power lines or noise as structural components. Such errors can lead to unnecessary inspections, wasting valuable resources and time. For example, an FP on a power line might trigger a maintenance alert for a non-existent fault.
    \item \textbf{False Negatives (FN)}: arise when the model fails to identify points that belong to a class. FN errors are particularly critical in power grid inspections, as they can result in undetected faults or risks. Missing a section of a power line, a damaged tower, or vegetation intrusion could lead to severe consequences, including power outages, structural failures, or wildfire risks.
\end{itemize}

Thus, these categories carry significant operational implications: effective power grid inspections demand models that minimize FP errors to improve operational efficiency and FN errors to ensure the safety and reliability of the grid.

\subsubsection{Intersection over Union}
The confusion matrix provides the foundation for calculating evaluation metrics such as the mean intersection-over-union (mIoU), which measures the accuracy of predictions for all semantic classes. mIoU is defined as:
\begin{equation}
\text{mIoU} = \frac{1}{C} \sum_{c=1}^{C} \frac{TP_c}{TP_c + FP_c + FN_c},
\end{equation}
where $C$ is the total number of classes, and $TP_c$, $FP_c$, and $FN_c$ correspond to the true positives, false positives, and false negatives for class $c$, respectively.

\subsubsection{F$_\beta$ Score}
The F$_\beta$ score is a performance metric used to evaluate the balance between precision and recall. It is defined as follows:
\begin{equation}
F_\beta = (1 + \beta^2) \cdot \frac{\text{Precision} \cdot \text{Recall}}{(\beta^2 \cdot \text{Precision}) + \text{Recall}},
\end{equation}
Here, Precision measures the proportion of correctly identified positive points out of all points predicted as positive, and Recall measures the proportion of correctly identified positive points out of all actual positive points. The parameter $\beta$ determines the relative weight given to recall versus precision: $\beta = 1$ (F$_1$ score) balances precision and recall equally; $\beta > 1$ places greater emphasis on recall, prioritizing the minimization of false negatives; $\beta < 1$ gives more importance to precision, focusing on reducing false positives.

In power grid inspections, the choice of $\beta$ depends on the task's operational priorities: F$_1$ Score is suited for scenarios where both false positives and false negatives have similar costs. For example, detecting vegetation risks might require a balanced approach to ensure thorough coverage without overwhelming operators with false alerts. 
The F$_{0.5}$ score becomes more appropriate when, for instance, excessive false alerts disrupt maintenance schedules or lead to unnecessary resource allocation, prioritizing precision is crucial.
Conversely, the F$_2$ score is ideal for situations where missing a potential failure, such as an undetected fault in a power line can lead to severe consequences like outages or fire hazards.
Therefore, the choice of $\beta$ aligns model performance with inspection goals. 
%

By using the F$_\beta$ score as an alternative to mIoU, operators can tailor model deployment: high F$_2$ scores suit safety-critical tasks, high F$_{0.5}$ scores optimize cost-sensitive scenarios, and F$_1$ balances efficiency and reliability. 
This adaptability enhances both safety and efficiency in grid maintenance.

\subsection{Results and Discussion}

\subsubsection{3D Semantic Segmentation on the TS40K Dataset}

\begin{table}[ht]
\centering
\caption{Benchmark results of 3D semantic segmentation baselines on the TS40K test set with a weighting scheme applied to prioritize underrepresented classes, namely supporting towers and power lines. We report mean IoU (mIoU) and per-class IoU scores.}
\label{tab:ts40k_class_weights}
\resizebox{\columnwidth}{!}{
\begin{tabular}{l|c|cccccc}
\hline
\textbf{Model} & \textbf{Mean IoU} & \textbf{Noise} & \textbf{Ground} & \textbf{Low Veg} & \textbf{Med Veg} & \textbf{Tower} & \textbf{Power Line} \\
\hline
PTV3~\cite{wu2023ptv3,pointcept2023} & \textbf{0.5858} & --- & 0.7613 & 0.6411 & \textbf{0.4837} & \textbf{0.5112} & 0.5316 \\
PTV2~\cite{wu2022point,pointcept2023} & 0.5712 & --- & \textbf{0.8065} & \textbf{0.6729} & 0.4639 & 0.4300 & 0.4829 \\
PTV1~\cite{zhao2021point,pointcept2023} & 0.5607 & --- & 0.7808 & 0.6228 & 0.4293 & 0.4736 & 0.4970 \\
KPConv~\cite{thomas2019kpconv} & 0.4386 & --- & 0.6495 & 0.3931 & 0.3187 & 0.3249 & 0.5069 \\
PointNet~\cite{qi2017pointnet} & 0.4267 & --- & 0.5866 & 0.5350 & 0.1643 & 0.1153 & 0.7324 \\
PointNet++~\cite{qi2017pointnet++} & 0.5061 & --- & 0.6690 & 0.5961 & 0.1796 & 0.2886 & \textbf{0.7972} \\
RandLaNet~\cite{hu2020randla} & 0.0652 & --- & 0.1673 & 0.0000 & 0.1589 & 0.0000 & 0.0000 \\
\hline
\end{tabular}}
\end{table}

\paragraph{Focusing on power grid elements during training}
Table~\ref{tab:ts40k_class_weights} presents the results of evaluating state-of-the-art models on the TS40K dataset, using a weighting scheme to prioritize underrepresented classes, particularly towers and power lines. Transformer-based models, such as PTV1~\cite{zhao2021point}, PTV2~\cite{wu2022point}, and PTV3~\cite{wu2023ptv3}, lead with the highest mean IoU (mIoU) scores of 56.02\%, 57.12\%, and 58.58\%, respectively. In contrast, PointNet~\cite{qi2017pointnet} and PointNet++~\cite{qi2017pointnet++} show a notable improvement in power line detection, with PointNet++ achieving a 26.56\% higher power line IoU compared to PTV3. However, these models fall short in detecting supporting towers, with IoU scores of just 11.53\% and 28.86\%, respectively. Despite their strengths in power line detection, the overall performance is not adequate for high-risk applications such as power grid inspection, where accurate identification of critical elements like towers and power lines is essential.

\begin{table}[ht]
\centering
\caption{Benchmark results of 3D semantic segmentation baselines on the TS40K test set with weighting scheme and detecting the noise class during training. All baselines showed improved performance for power grid elements, as a significant portion of the noise 3D points was misclassified as towers or power lines.}
\label{tab:ts40k_noise}
\resizebox{\columnwidth}{!}{
\begin{tabular}{l|c|cccccc}
\hline
\textbf{Model} & \textbf{Mean IoU} & \textbf{Noise} & \textbf{Ground} & \textbf{Low Veg} & \textbf{Med Veg} & \textbf{Tower} & \textbf{Power Line} \\
\hline
PTV3~\cite{wu2023ptv3,pointcept2023} & 0.6355 & 0.5923 & 0.7077 & 0.5047 & 0.4386 & \textbf{0.6142} & \textbf{0.9553} \\
PTV2~\cite{wu2022point,pointcept2023} & \textbf{0.6829} & 0.6116 & \textbf{0.8013} & \textbf{0.6817} & \textbf{0.5139} & 0.5448 & 0.9443 \\
PTV1~\cite{zhao2021point,pointcept2023} & 0.6490 & 0.5750 & 0.7733 & 0.6034 & 0.4651 & 0.5419 & 0.9354 \\
KPConv~\cite{thomas2019kpconv}    & 0.5277 & 0.5702 & 0.6475 & 0.3712 & 0.3463 & 0.3736 & 0.8999 \\
PointNet++~\cite{qi2017pointnet++} & 0.4599 & 0.5927 & 0.5999 & 0.5436 & 0.1455 & 0.2261 & 0.7841 \\
PointNet~\cite{qi2017pointnet} & 0.3001 & 0.4936 & 0.5452 & 0.4600 & 0.1423 & 0.0000 & 0.3528 \\
RandLaNet~\cite{hu2020randla} & 0.0650 & 0.0791 & 0.0000 & 0.0000 & 0.2158 & 0.0000 & 0.1092 \\
\hline
\end{tabular}}
\end{table}

\paragraph{Noise detection in power grid segmentation}
Upon analyzing the confusion matrix from the models in Table~\ref{tab:ts40k_class_weights}, we observe that noise points are frequently misclassified as power grid elements, particularly power lines, likely due to their similar height. Noise points in the TS40K dataset primarily originate from LiDAR sensor artifacts during inspections, with harsh weather and lightning conditions further contributing to their density. While it is typically avoided to explicitly detect noise in segmentation tasks, noise points have a higher point density than power grid elements in TS40K. Thus, incorporating noise point detection during model training can enhance the overall segmentation of power grid elements.
As shown in Table~\ref{tab:ts40k_noise}, the inclusion of noise detection results in a notable improvement in the detection of power grid elements, specifically towers and power lines. Transformer-based models~\cite{zhao2021point,wu2022point,wu2023ptv3} show the most significant improvement, with PTV3~\cite{wu2023ptv3} achieving an IoU of 95.53\% for power line detection and PTV2~\cite{wu2022point} achieves a mIoU of 68.29\%, an increase of 11.17\% to its performance in Table~\ref{tab:ts40k_class_weights}.

\begin{table}[ht]
\centering
\caption{Benchmark results of 3D semantic segmentation baselines on the TS40K test set, incorporating normal vectors for every 3D point. The inclusion of normal vectors resulted in performance similar to that reported in~\ref{tab:ts40k_noise}, indicating that normal vectors do not contribute to improved segmentation of the TS40K dataset.}
\label{tab:ts40k_normals}
\resizebox{\columnwidth}{!}{
\begin{tabular}{l|c|cccccc}
\hline
\textbf{Model} & \textbf{Mean IoU} & \textbf{Noise} & \textbf{Ground} & \textbf{Low Veg} & \textbf{Med Veg} & \textbf{Tower} & \textbf{Power Line} \\
\hline
PTV3~\cite{wu2023ptv3,pointcept2023} & 0.6400 & 0.5921 & 0.7055 & 0.5318 & 0.4427 & \textbf{0.6116} & \textbf{0.9565} \\
PTV2~\cite{wu2022point,pointcept2023} & 0.6677 & \textbf{0.6262} & 0.7820 & 0.6166 & 0.4728 & 0.5636 & 0.9451 \\
PTV1~\cite{zhao2021point,pointcept2023} & \textbf{0.6775} & 0.6127 & \textbf{0.7914} & \textbf{0.6505} & \textbf{0.4969} & 0.5669 & 0.9468 \\
KPConv~\cite{thomas2019kpconv}    & 0.5595 & 0.5648 & 0.6706 & 0.4208 & 0.3689 & \textcolor{red}{0.4208} & 0.9165 \\
PointNet++~\cite{qi2017pointnet++} & 0.4832 & 0.6038 & 0.6017 & 0.5514 & 0.1604 & \textcolor{red}{0.3002} & 0.8023 \\
PointNet~\cite{qi2017pointnet} & 0.3984 & 0.5183 & 0.5654 & 0.4832 & 0.2039 & 0.0934 & \textcolor{red}{0.5262} \\
RandLaNet~\cite{hu2020randla} & 0.0817 & 0.0923 & 0.0031 & 0.0465 & 0.1342 & 0.0105 & 0.1046 \\
\hline
\end{tabular}}
\end{table}

\paragraph{Including normal vectors}
Normal vectors, which represent the orientation of surface elements, have been widely used in literature to enhance 3D point cloud segmentation. They provide geometric context that helps distinguish between different surface types, such as flat (ground) and vertical (towers), thereby improving object boundary detection. 
Thus, normal vectors were added to the training setting of Table~\ref{tab:ts40k_noise}. However, upon analyzing the results from Table~\ref{tab:ts40k_normals}, the inclusion of normal vectors does not significantly improve the segmentation accuracy for the TS40K dataset when compared to the performance in Table~\ref{tab:ts40k_noise}. 
Despite the inclusion of this additional feature, the performance of transformer-based models~\cite{zhao2021point,wu2022point,wu2023ptv3} remains largely similar, whereas other baselines, such as KPConv~\cite{thomas2019kpconv} and PointNet++~\cite{qi2017pointnet++} show moderate improvement (highlighted in red in Table~\ref{tab:ts40k_normals}).
While normal vectors provide some improvement in certain categories, they do not substantially enhance the overall mean IoU or contribute to a significant boost in detecting power grid elements compared to the noise detection strategy.

\begin{table}[ht]
\centering
\caption{Benchmark results of 3D semantic segmentation baselines on the TS40K test set, excluding the ground class during training. Transmission network operators often implement routines to automatically remove ground 3D points before analyzing power grids in 3D point cloud format. By incorporating this routine into the training of state-of-the-art methods, we observed an improvement in the detection of power grid elements.}
\label{tab:ts40k_no_ground}
\resizebox{\columnwidth}{!}{
\begin{tabular}{l|c|cccccc}
\hline
\textbf{Model} & \textbf{Mean IoU} & \textbf{Noise} & \textbf{Ground} & \textbf{Low Veg} & \textbf{Med Veg} & \textbf{Tower} & \textbf{Power Line} \\
\hline
PTV3~\cite{wu2023ptv3,pointcept2023} & 0.6746 & 0.6467 & --- & 0.6424 & 0.4708 & \textbf{0.6505} & \textbf{0.9625} \\
PTV2~\cite{wu2022point,pointcept2023} & \textbf{0.7286} & \textbf{0.7089} & --- & \textbf{0.7686} & \textbf{0.5901} & 0.6169 & 0.9583 \\
PTV1~\cite{zhao2021point,pointcept2023} & 0.6789 & 0.5825 & --- & 0.7428 & 0.5469 & 0.5715 & 0.9507 \\
KPConv~\cite{thomas2019kpconv}    & 0.4731 & 0.6144 & --- & 0.5903 & 0.4213 & 0.4273 & 0.9264 \\
PointNet++~\cite{qi2017pointnet++} & 0.4272 & 0.6135 & --- & 0.7608 & 0.2794 & 0.2588 & 0.8370 \\
PointNet~\cite{qi2017pointnet} & 0.3668 & 0.4932 & --- & 0.4574 & 0.1978 & 0.0857 & 0.4569 \\
RandLaNet~\cite{hu2020randla} & 0.0725 & 0.0763 & --- & 0.0341 & 0.1926 & 0.0000 & 0.1204 \\
\hline
\end{tabular}}
\end{table}

\begin{figure}[t]
\centering
\begin{subfigure}[b]{0.44\columnwidth}
  \includegraphics[width=\linewidth]{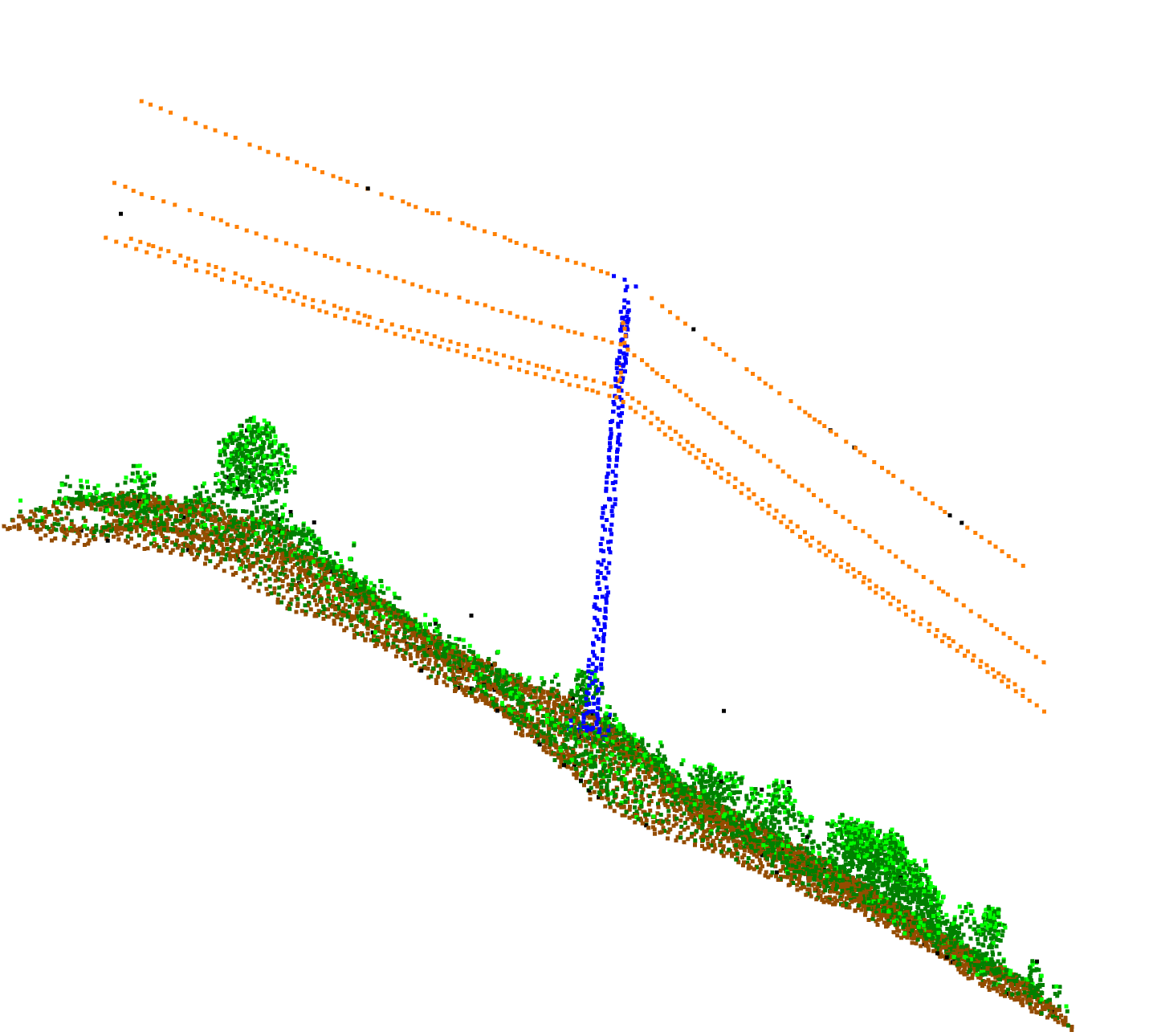}
\end{subfigure}
\hfill
\begin{subfigure}[b]{0.44\columnwidth}
  \includegraphics[width=\linewidth]{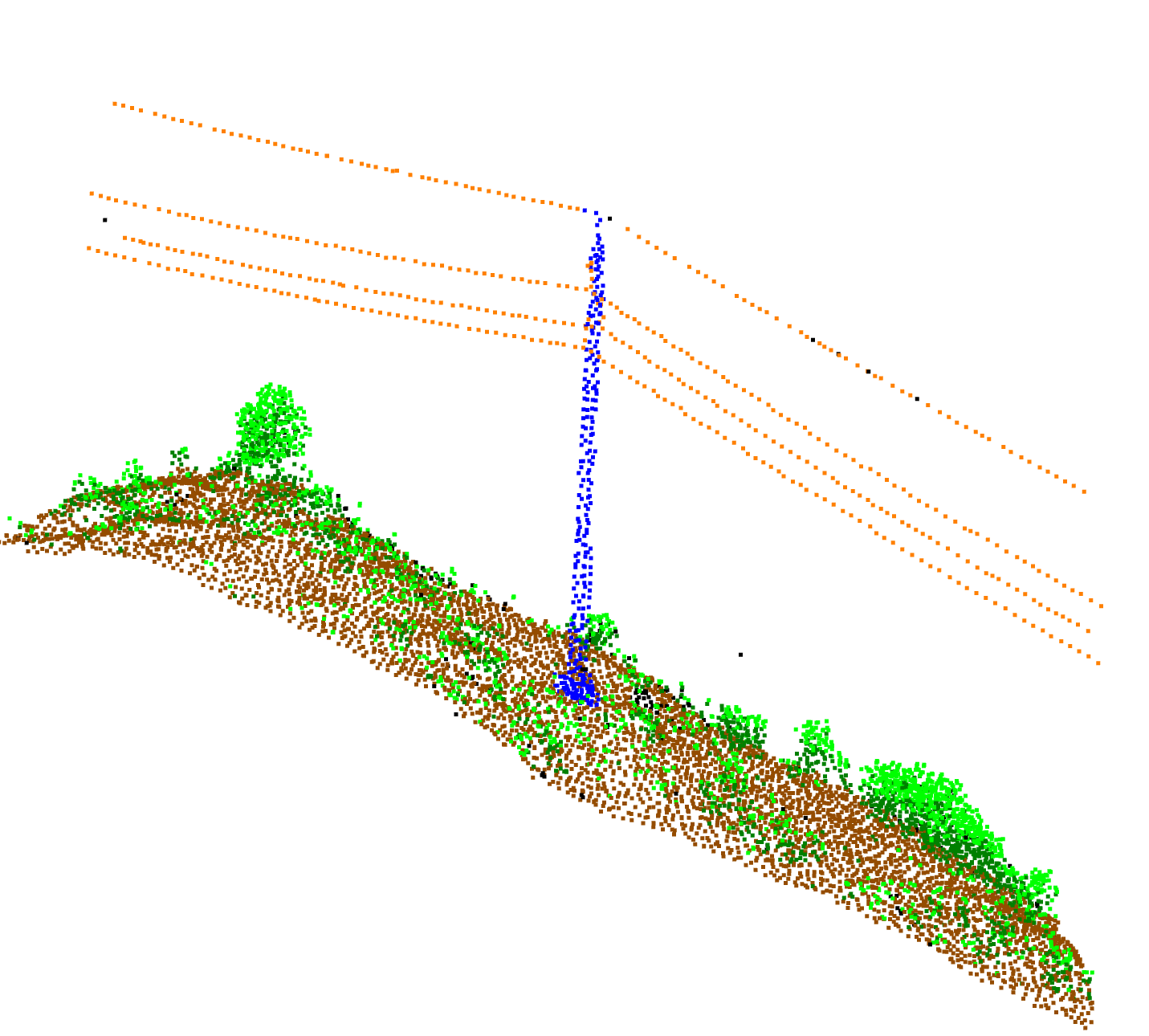}
\end{subfigure}
\\
\centering
\begin{subfigure}[b]{0.44\columnwidth}
  \includegraphics[width=\linewidth]{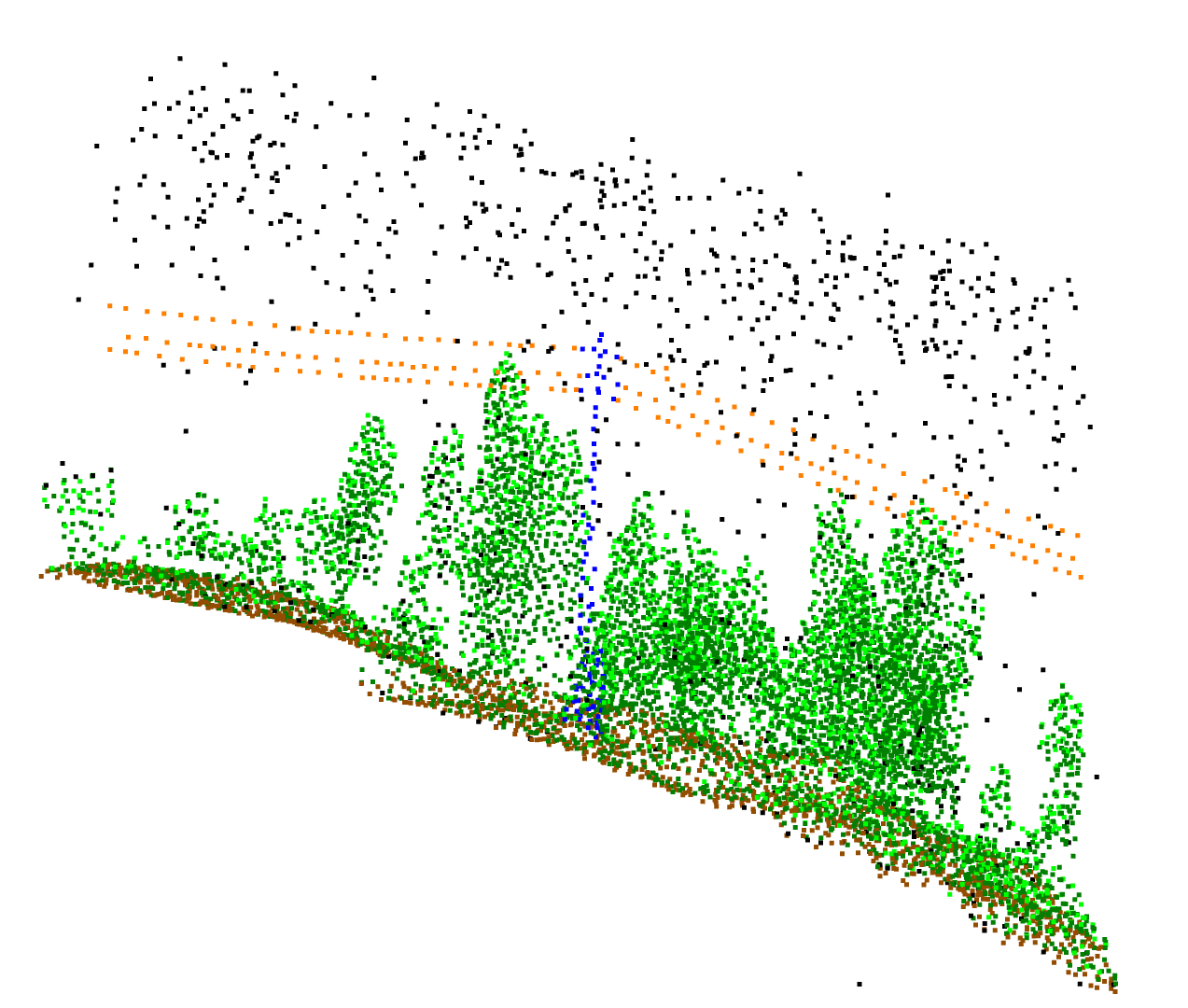}
  \caption{Ground Truth}
\end{subfigure}
\hfill
\begin{subfigure}[b]{0.44\columnwidth}
  \includegraphics[width=\linewidth]{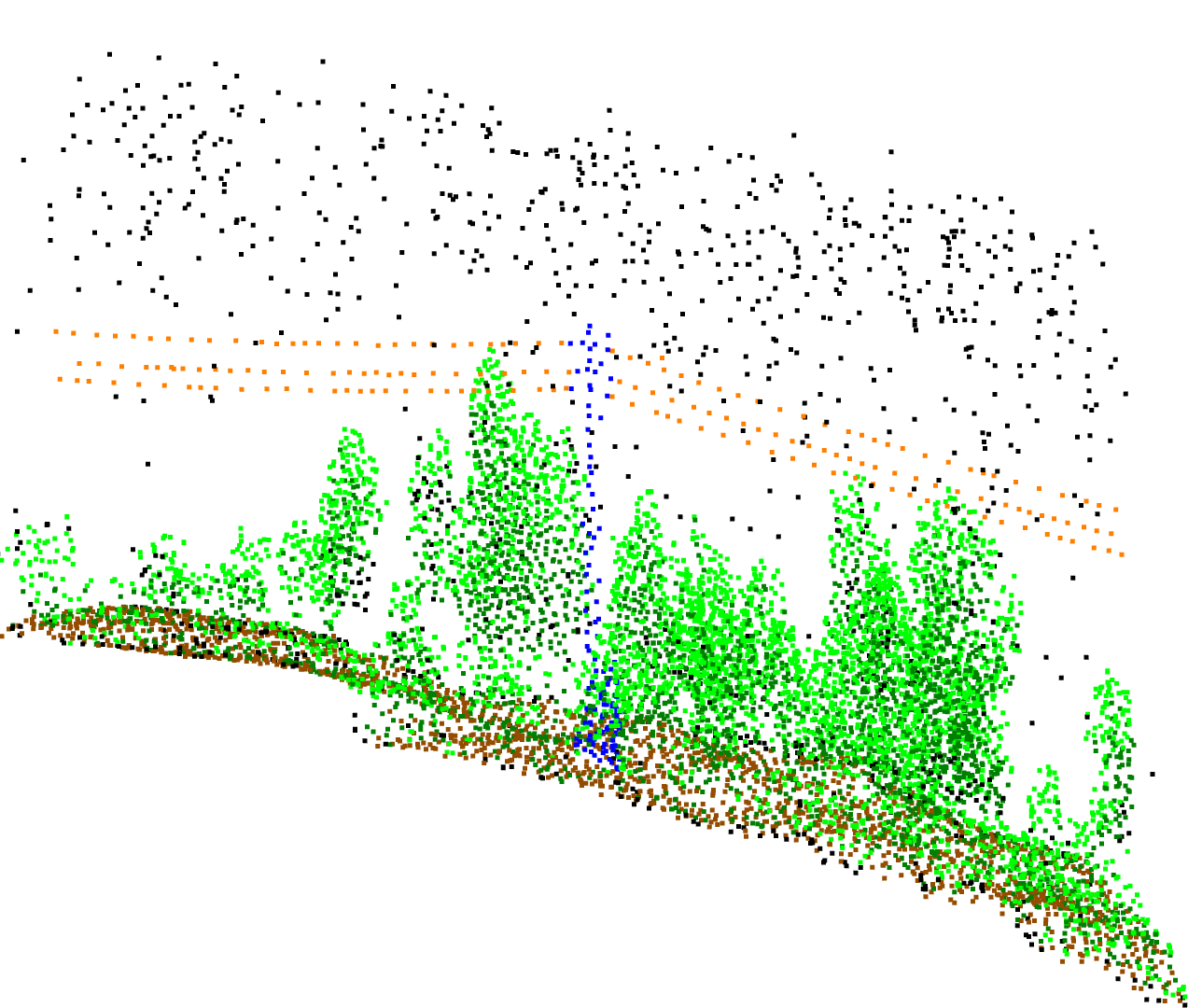}
  \caption{PTV3 Prediction}
\end{subfigure}
\\
\includegraphics[width=1\columnwidth]{ts40k_teaser/legend_classes.png}
\caption{Qualitative results showcasing the performance of Point Transformer V3 (PTV3)~\cite{wu2023ptv3} on the TS40K dataset. While PTV3 is not the highest mIoU performing model, it consistently achieves the highest segmentation performance in crucial inspection elements, namely supporting towers and power lines. Thus, it is particularly well-suited for tasks prioritizing the accurate detection of these components in power grid inspections.
}
\label{fig:ptv3-ts40k}
\end{figure}

\paragraph{Disregarding ground points}
Transmission network inspectors often leverage empirically tested heuristics to streamline the annotation of 3D point clouds in LiDAR-based inspections. Less critical elements, such as ground and low vegetation, are typically annotated automatically, while key components like supporting towers, power lines, and medium vegetation receive manual annotations for higher precision.
Ground points account for a substantial 55.28\% of the TS40K dataset~\cite{lavado2024ts40k}. By excluding these points through heuristic-based filtering, baseline models can focus computational resources on more critical elements, enhancing segmentation accuracy. Although of lower priority, low vegetation remains a relevant class in power grid inspection due to its potential to pose collision risks in certain scenarios.
Table~\ref{tab:ts40k_no_ground} highlights the performance improvements of state-of-the-art models when ground points are excluded. Transformer-based methods~\cite{zhao2021point,wu2022point,wu2023ptv3} demonstrate notable gains, with PTV3~\cite{wu2023ptv3} achieving the highest IoU for power lines (96.25\%) and supporting towers (65.05\%) across all experiments.

\subsection{3D Semantic Segmentation on the TS-RGB Dataset}

\begin{figure}[ht]
    \centering
    \includegraphics[width=\columnwidth]{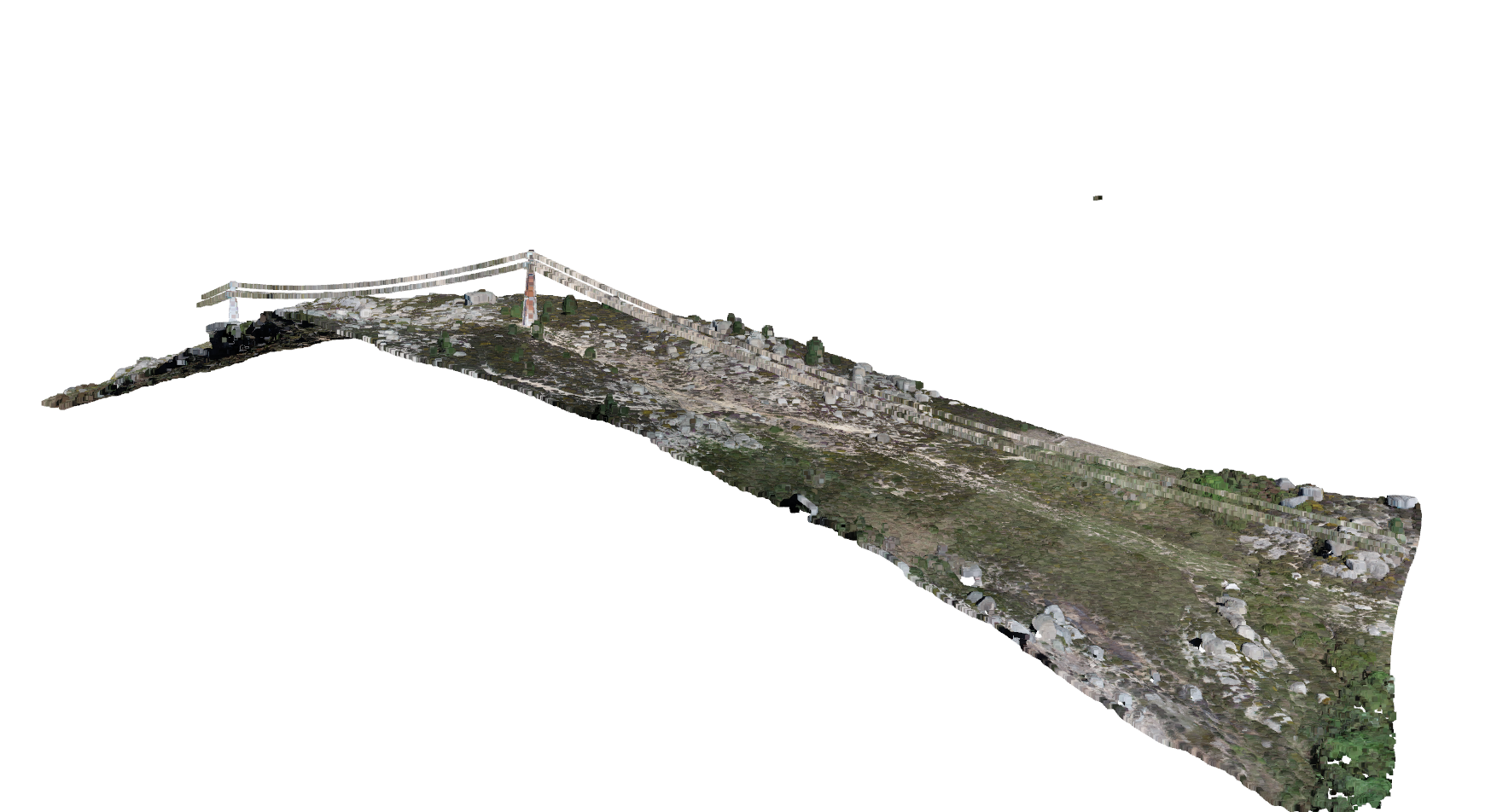}
    \caption{Visualization of the TS-RGB dataset. TS-RGB is an augmented version of the TS40K dataset, incorporating RGB channels to improve 3D semantic segmentation in power grid environments. Covering approximately 8,000 kilometers of transmission network, it includes over 1,295 million points collected using LiDAR sensors. Ground points are automatically removed by heuristics and annotations exclude differentiation between low and medium vegetation.}
    \label{fig:ts_rgb_dataset}
\end{figure}

TS-RGB is an reduced version of the TS40K dataset, augmented with RGB color channels. This dataset covers approximately 8,000 kilometers of power grid infrastructure and contains over 1,295 million points, as shown in~\ref{fig:ts_rgb_dataset}.
While TS-RGB shares the same transmission network as TS40K, it was collected using a more advanced LiDAR sensor, resulting in significantly higher point density. Unlike TS40K, the differentiation between low and medium vegetation is not accounted for in annotation practices by maintenance personnel. Additionally, ground-labeled points are automatically removed by well tested heuristics in the TS-RGB dataset.
The addition of RGB channels in recent 3D benchmarks has been shown to improve the performance of state-of-the-art methods in 3D semantic segmentation. Motivated by this, we investigate the impact of incorporating RGB information to enhance segmentation accuracy for power grid inspection tasks.

\begin{table}[ht]
    \centering
    \caption{Benchmark results of 3D semantic segmentation baselines on the TS-RGB test set, trained without RGB channels using a training strategy consistent with that of the TS40K dataset. This uniform setup ensures a fair comparison of baseline model performance across both datasets. Reported metrics include the mean IoU (mIoU) and per-class IoU scores.}
    \resizebox{\columnwidth}{!}{
    \begin{tabular}{l|c|ccccc}
        \hline
        \textbf{Model} & \textbf{Mean IoU} & \textbf{Noise} & \textbf{Vegetation} & \textbf{Tower} & \textbf{Power Line} \\
        \hline
        PTV3~\cite{wu2023ptv3,pointcept2023}         & 0.5915 & 0.2924 & 0.9148 & 0.3814 & 0.7772 \\
        PTV2~\cite{wu2022point,pointcept2023}        & \textbf{0.5965} & 0.3107 & 0.9175 & \textbf{0.3845} & 0.7734 \\
        PTV1~\cite{zhao2021point,pointcept2023}      & 0.5828 & 0.3125 &\textbf{0.9277}& 0.3136 & \textbf{0.7775} \\
        KPConv~\cite{thomas2019kpconv}               & 0.4912 & 0.1983 & 0.8709 & 0.2194 & 0.6834 \\
        PointNet++~\cite{qi2017pointnet++}           & 0.4302 & \textbf{0.5158} & 0.1489 & 0.2803 & 0.6204 \\
        PointNet~\cite{qi2017pointnet}               & 0.3659 & 0.4557 & 0.1763 & 0.0859 & 0.4080 \\
        RandLaNet~\cite{hu2020randla}                & 0.0701 & 0.0857 & 0.1162 & 0.0128 & 0.0927 \\
        \hline
    \end{tabular}
    }
    \label{tab:labelec_xyz}
\end{table}

\begin{figure}[t]
\centering
\begin{subfigure}[b]{0.44\columnwidth}
  \includegraphics[width=\linewidth]{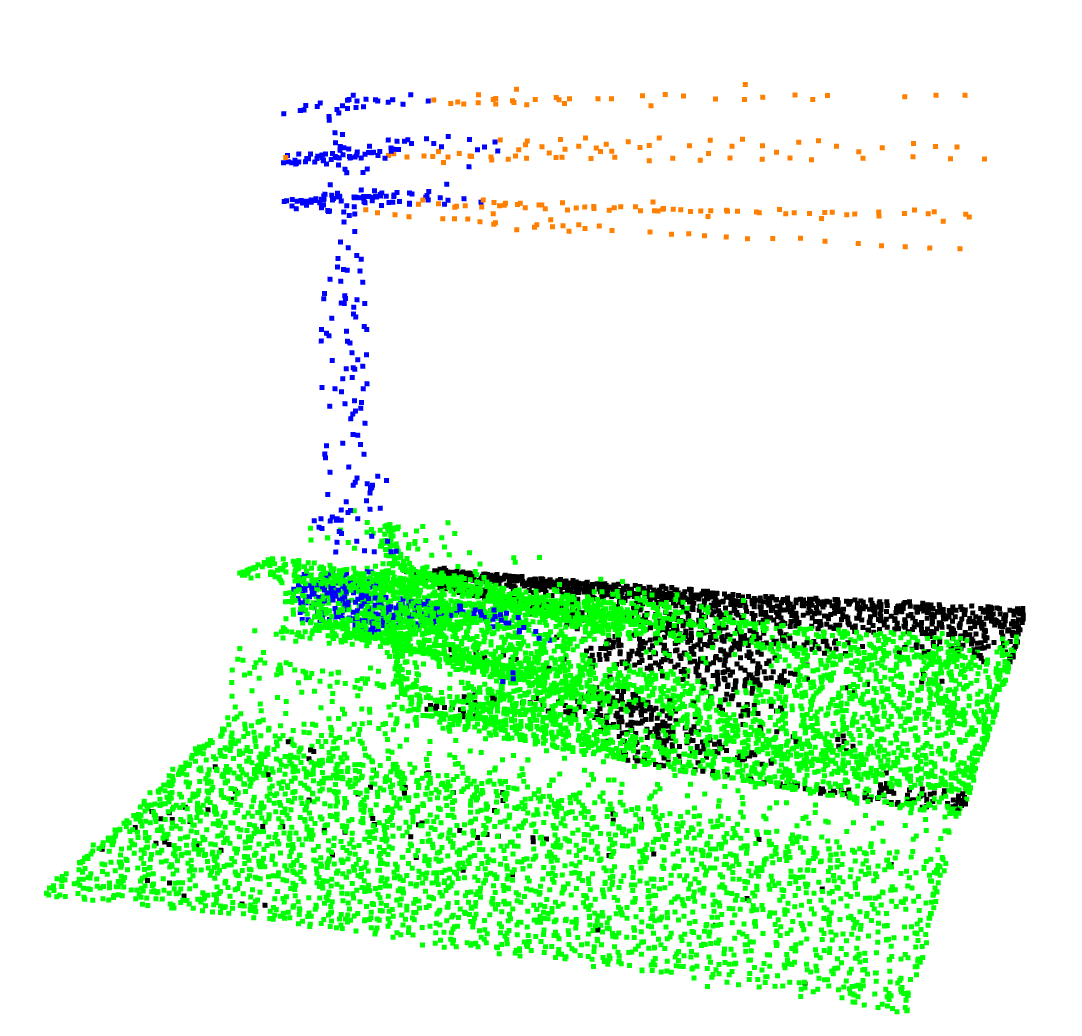}
\end{subfigure}
\hfill
\begin{subfigure}[b]{0.44\columnwidth}
  \includegraphics[width=\linewidth]{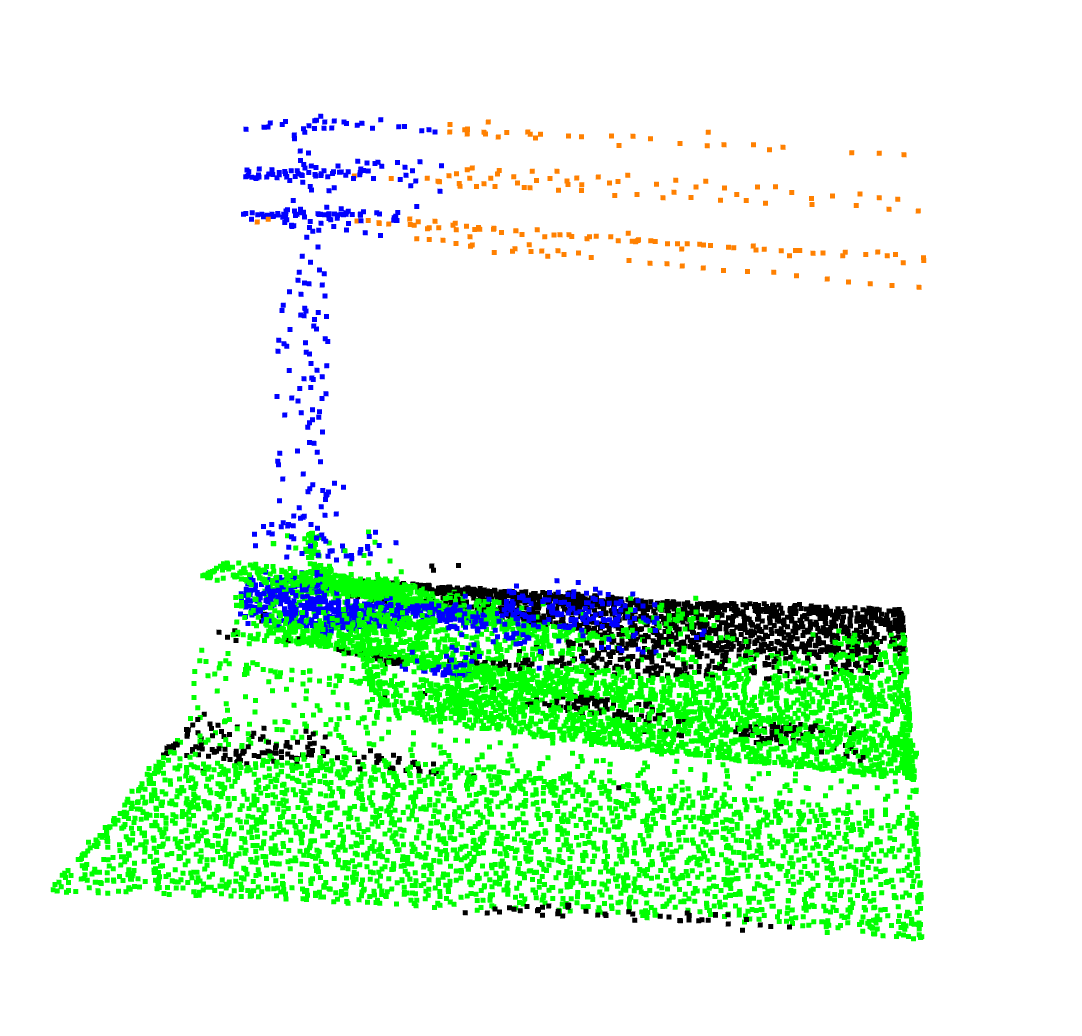}
\end{subfigure}
\\
\centering
\begin{subfigure}[b]{0.44\columnwidth}
  \includegraphics[width=\linewidth]{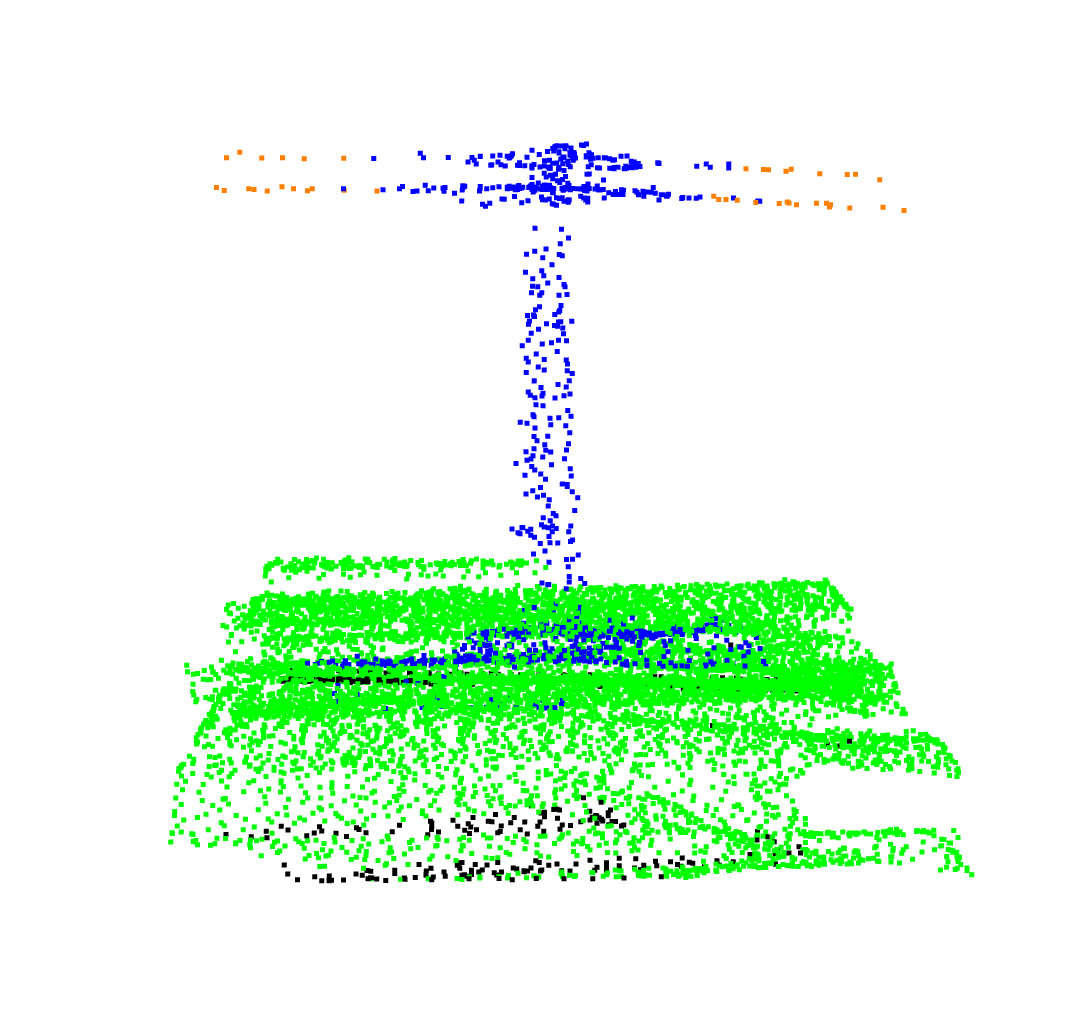}
  \caption{Ground Truth}
\end{subfigure}
\hfill
\begin{subfigure}[b]{0.44\columnwidth}
  \includegraphics[width=\linewidth]{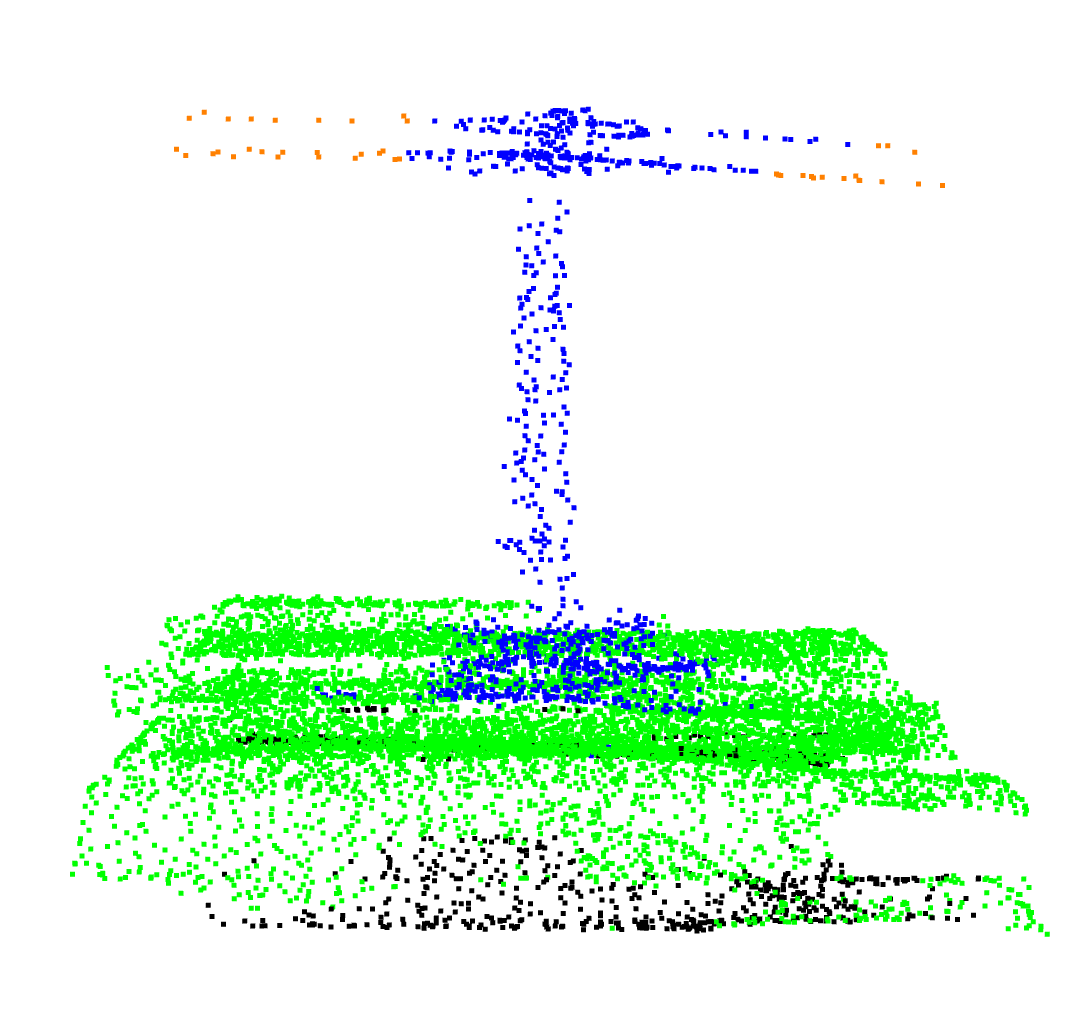}
  \caption{PTV2 Prediction}
\end{subfigure}
\\
\includegraphics[width=1\columnwidth]{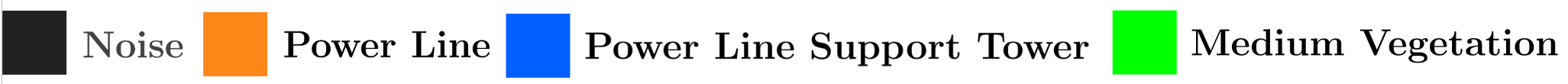}
\caption{Qualitative results illustrating the performance of Point Transformer V2 (PTV2)~\cite{wu2022point} on the TS-RGB dataset, utilizing point coordinates as the sole input features. Table~\ref{tab:labelec_xyz} demonstrates that PTV2 achieves the highest mean IoU and outperforms all other models in tower segmentation across all experiments in this configuration. These results underscore PTV2's effectiveness and reliability for semantic segmentation tasks, particularly in high-stakes applications such as power grid inspections.
}
\label{fig:ptv3-tsrgb}
\end{figure}

\paragraph{Analysis of TS-RGB results using only coordinates as features}
The results in Table~\ref{tab:labelec_xyz} showcase the performance of state-of-the-art 3D semantic segmentation models on the TS-RGB dataset when trained using only point coordinates, omitting the RGB information. This evaluation enables a direct comparison to previous benchmarks, such as TS40K, under consistent conditions.
Transformer-based models~\cite{zhao2021point,wu2022point,wu2023ptv3} demonstrate superior performance, achieving the highest mean IoU scores among all baselines, with PTV2~\cite{wu2022point} slightly outperforming others at 59.65\%. This consistent trend highlights the effectiveness of attention mechanisms in capturing spatial relationships, even when additional features like color are absent. Conversely, simpler architectures like PointNet~\cite{qi2017pointnet} and PointNet++~\cite{qi2017pointnet++} show lower mIoU, emphasizing the limitations of these methods in processing complex spatial patterns in datasets with high variability.
The per-class analysis highlights notable strengths and weaknesses among the methods. Transformer-based models excel in segmenting vegetation, with PTV3~\cite{wu2023ptv3} achieving a remarkable IoU of 91.48\% for this class. However, their performance on the tower and power line classes is comparatively weaker than on TS40K. This disparity can be attributed to the dominance of dense vegetation (92\% of points) and the sparse representation of power grid elements (approximately 1.4\%), which complicate segmentation. Noise points remain a persistent challenge across all models, with low IoU scores reflecting the difficulty of distinguishing artifacts with limited geometric cues.

\begin{table}[ht]
    \centering
    \caption{Benchmark results of 3D semantic segmentation baselines on the TS-RGB test set with RGB channels.}
    \resizebox{\columnwidth}{!}{
    \begin{tabular}{l|c|ccccc}
        \hline
        \textbf{Model} & \textbf{Mean IoU} & \textbf{Noise} & \textbf{Vegetation} & \textbf{Tower} & \textbf{Power Line} \\
        \hline
        PTV3~\cite{wu2023ptv3,pointcept2023}      & \textbf{0.5934} & 0.2951 & \textbf{0.9217} & \textbf{0.3821} & \textbf{0.7745}\\
        PTV2~\cite{wu2022point,pointcept2023}     & 0.5109 & 0.2678 & 0.9088 & 0.1397 & 0.7274 \\
        PTV1~\cite{zhao2021point,pointcept2023}   & 0.5227 & 0.2884 & 0.9148 & 0.1845 & 0.7030 \\
        KPConv~\cite{thomas2019kpconv}            & 0.5081 & 0.2035 & 0.8931 & 0.2285 & 0.7073 \\
        PointNet++~\cite{qi2017pointnet++}        & 0.4301 & \textbf{0.5378} & 0.1431 & 0.2684 & 0.6160 \\
        PointNet~\cite{qi2017pointnet}            & 0.3542 & 0.4615 & 0.1816 & 0.0831 & 0.4047 \\
        RandLaNet~\cite{hu2020randla}             & 0.0725 & 0.0820 & 0.1184 & 0.0098 & 0.0932 \\
        \hline
    \end{tabular}
    }
    \label{tab:labelec_rgb}
\end{table}

\paragraph{Analysis of results on TS-RGB with color channels}
The inclusion of RGB channels provides limited improvements in segmentation performance. While PTV3~\cite{wu2023ptv3} achieves a slightly higher mean IoU (59.34\% vs. 59.15\%) and better vegetation IoU (92.17\% vs. 91.48\%), gains in other classes, such as towers and noise, remain marginal. PTV2~\cite{wu2022point} and PTV1~\cite{zhao2021point} exhibit minor fluctuations in performance, with some classes showing slight declines compared to results without RGB. Traditional methods like KPConv~\cite{thomas2019kpconv} show negligible improvements or even reduced performance in certain cases. These findings suggest that while RGB enhances segmentation of dominant classes like vegetation, it provides limited benefits for harder-to-distinguish elements such as towers and noise.
An analysis of the RGB channels in the TS-RGB dataset reveals minimal color differentiation between vegetation and power grid elements (as shown in~\ref{fig:ts_rgb_dataset}). Vegetation and power grid components are often either similarly dark green with subtle shade variations, or the power grid appears gray while vegetation remains dark green. This lack of distinct color contrast may account for the limited improvements observed in segmentation performance on the TS-RGB dataset.

\begin{table}[ht]
    \centering
    \caption{Benchmark results of 3D semantic segmentation baselines on the TS-RGB test set with RGB and normal vectors as additional features.}
    \resizebox{\columnwidth}{!}{
    \begin{tabular}{l|c|ccccc}
        \hline
        \textbf{Model} & \textbf{Mean IoU} & \textbf{Noise} & \textbf{Vegetation} & \textbf{Tower} & \textbf{Power Line} \\
        \hline
        PTV3~\cite{wu2023ptv3,pointcept2023}         & 0.5797 & 0.2902 & \textbf{0.9271} & 0.3356 & 0.7658 \\
        PTV2~\cite{wu2022point,pointcept2023}        & 0.5425 & 0.2865 & 0.9142 & 0.2365 & 0.7329 \\
        PTV1~\cite{zhao2021point,pointcept2023}      & \textbf{0.5927} & 0.3147 & 0.9157 & \textbf{0.3499} & \textbf{0.7906} \\
        KPConv~\cite{thomas2019kpconv}               & 0.5245 & 0.2210 & 0.9123 & 0.3075 & 0.6571 \\
        PointNet++~\cite{qi2017pointnet++}           & 0.4103 & \textbf{0.5532} & 0.1350 & 0.2901 & 0.6025 \\
        PointNet~\cite{qi2017pointnet}               & 0.3702 & 0.4508 & 0.1705 & 0.0952 & 0.4158 \\
        RandLaNet~\cite{hu2020randla}                & 0.0805 & 0.0789 & 0.1253 & 0.0154 & 0.0886 \\
        \hline
    \end{tabular}
    }
    \label{tab:labelec_rgb_normal}
\end{table}

\paragraph{Analysis of results on TS-RGB with color and normal vectors}
When integrating both RGB information and normal vectors, models exhibit mixed performance compared to earlier configurations. PTV1~\cite{zhao2021point} stands out as the best performer, achieving the highest mean IoU at 59.27\%, outperforming its successors PTV2~\cite{wu2022point} and PTV3~\cite{wu2023ptv3}. However, the IoU for supporting towers shows a notable decline, dropping by 5\% compared to both the RGB-only and coordinates-only setups.
experiences a drop in power line IoU (76.58\%) compared to the RGB-only configuration. The inclusion of normal vectors does not yield consistent improvements across all models. For instance, KPConv~\cite{thomas2019kpconv} and PointNet++~\cite{qi2017pointnet++} exhibit reduced mean IoU and weaker performance in tower segmentation. Overall, while the addition of normal vectors benefits certain models, particularly in power line segmentation, it provides limited advantages for vegetation segmentation and noise detection.

\begin{table}[ht]
    \centering
    \caption{Benchmark results of 3D semantic segmentation baselines on the TS-RGB test set using  only normal vectors as input features.}
    \resizebox{\columnwidth}{!}{
    \begin{tabular}{l|c|ccccc}
        \hline
        \textbf{Model} & \textbf{Mean IoU} & \textbf{Noise} & \textbf{Vegetation} & \textbf{Tower} & \textbf{Power Line} \\
        \hline
        PTV3~\cite{wu2023ptv3,pointcept2023}         & 0.5846 & 0.2977 & 0.9213 & 0.3375 & \textbf{0.7819} \\
        PTV2~\cite{wu2022point,pointcept2023}        & \textbf{0.5905} & 0.2971 & 0.9209 & \textbf{0.3664} & 0.7778 \\
        PTV1~\cite{zhao2021point,pointcept2023}      & 0.5666 & 0.3227 & \textbf{0.9284} & 0.2570 & 0.7585 \\
        KPConv~\cite{thomas2019kpconv}               & 0.5271 & 0.2044 & 0.9032 & 0.3002 & 0.7007 \\
        PointNet++~\cite{qi2017pointnet++}           & 0.4456 & \textbf{0.5291} & 0.1504 & 0.2550 & 0.6357 \\
        PointNet~\cite{qi2017pointnet}               & 0.3487 & 0.4742 & 0.1902 & 0.0771 & 0.3974 \\
        RandLaNet~\cite{hu2020randla}                & 0.0652 & 0.0910 & 0.1103 & 0.0086 & 0.0981 \\
        \hline
    \end{tabular}
    }
    \label{tab:labelec_normal}
\end{table}

\paragraph{Analysis of results on TS-RGB solely with normal vectors}
In this experiment, where only normal vectors are used as input features, the models show varying degrees of performance. PTV2~\cite{wu2022point} achieves the highest mean IoU score (59.05\%), slightly outperforming PTV3~\cite{wu2023ptv3} (58.46\%). 
Models like KPConv~\cite{thomas2019kpconv} and PointNet++\cite{qi2017pointnet++} show relatively lower performance, particularly in tower and power line segmentation. RandLaNet~\cite{hu2020randla} struggles significantly, with a very low mean IoU (6.52\%) and poor performance across all classes. This indicates that while normal vectors offer improvements for some models, they still present challenges in tasks such as power line and tower segmentation.

\section{Inspection Tool for Power Grid Segmentation~\label{sec:inspection_tool}}

We propose inspection tool designed to leverage advanced 3D semantic segmentation models to automate power grid inspection tasks, with a specific focus on accurately identifying critical infrastructure components such as power lines and supporting towers. The tool is part of a broader strategy to improve the efficiency of power grid maintenance by integrating state-of-the-art machine learning techniques into the inspection workflow. The tool follows a systematic pipeline for processing and analyzing point clouds:

\begin{enumerate}
    \item \textbf{Point Cloud Partition:} The first step involves dividing the power grid point cloud into contiguous segments, each representing roughly a 50-meter stretch of the infrastructure. This segmentation ensures that each sample corresponds to a localized portion of the grid, making it computationally feasible to process large-scale point clouds while preserving the spatial continuity of the grid elements.
    \item \textbf{Preprocessing with Farthest Point Sampling (FPS):} Farthest Point Sampling (FPS) is a technique used to select a subset of points from a larger point cloud in such a way that the selected points are spread out as much as possible. The method iteratively selects the point that is farthest from the previously selected points, ensuring that the chosen points are well-distributed across the segment. For each segmented point cloud, FPS is applied to reduce the number of points to a consistent 100,000 points per segment. This process is crucial for standardizing the input size across point clouds with varying densities, while preserving the key geometric features needed for accurate segmentation. FPS ensures that the most representative points are retained, allowing the model to focus on relevant spatial information and improving computational efficiency.
    \item \textbf{Prediction:} The core step of the pipeline involves predicting labels for each point in the segmented cloud using a pretrained 3D semantic segmentation model. The model produces a softmax distribution for each point, which indicates the model's confidence in the predicted class for that point. These softmax scores reflect the model's certainty, with sharper distributions indicating high confidence and flatter distributions suggesting uncertainty.
    \item \textbf{Reconstruction and Label Propagation:} After segmentation, the predicted labels and 3D points are reconstructed back into the original point cloud. While this reconstruction results in fewer points than the initial point cloud, this reduction is acceptable for the inspection process, as the critical infrastructure components are still adequately represented. The reconstructed point cloud, now with predicted labels, serves as the output for further analysis and review.
\end{enumerate}

\subsection{Flagging Uncertain Predictions for Manual Review}
A key feature of the tool is its ability to identify and flag uncertain predictions. For each point in the segmented point cloud, the model computes a softmax distribution, which indicates the model’s confidence in its prediction. If the softmax distribution is not sharp, meaning the majority class is not clearly defined (i.e., the model's predictions are uncertain or ambiguous), the corresponding point is marked as "undecided." 
To handle this, a clustering algorithm is applied to group these undecided points, and if the number of undecided points in a segment exceeds a predefined safety threshold, the entire segment is flagged for manual inspection by maintenance personnel.
This flagging mechanism ensures that only segments where the model is uncertain are subject to manual review, thus reducing the overall workload for maintenance personnel while focusing their efforts on the most ambiguous or critical cases. This approach is particularly valuable for power grid inspections, where maintaining operational efficiency is crucial, and minimizing unnecessary manual interventions is a priority.

\subsection{Performance Requirements for Power Grid Inspection}
The choice of $\beta = 2$ for the F$_\beta$ score in power grid inspection tasks is rooted in the need to prioritize recall over precision. In this context, recall is more important because the potential consequences of missing a critical issue, such as a fault in the power grid infrastructure, can be severe. These issues can lead to power outages, safety hazards, or even catastrophic failures, which can have widespread implications. While IoU is more suited for model comparison in 3D semantic segmentation, F$_2$ is more appropriate to evaluate model application to power grid inspection. Thus, the models chosen are optimized to minimize false negatives, instances where important issues go undetected, ensuring that critical problems are flagged for attention.
While false positives, such as incorrectly identifying non-problematic elements (e.g., harmless vegetation) as issues, can cause operational inefficiencies and unnecessary maintenance, they are generally less harmful in the context of power grid inspections. False positives may result in additional inspections or resource allocation, but the consequences are typically more manageable compared to missing a critical fault in the infrastructure (i.e., false negatives).
Thus, by emphasizing recall with $\beta = 2$, the chosen model is designed to err on the side of caution, ensuring that as many potential issues as possible are flagged for further inspection. 
The emphasis on recall ensures that maintenance teams are alerted to any possible threats, even at the cost of an increased number of false positives, which can be addressed through subsequent verification and validation processes.

\begin{table}[ht]
\centering
\caption{F$_2$ scores for the performance of Point Transformer V3 (PTV3) on the TS40K and TS-RGB datasets. The results for TS40K demonstrate strong performance in power grid classes, while TS-RGB shows more variability, particularly in certain classes.}
\begin{tabular}{l|c|c}
\textbf{Class} & \textbf{TS40K F$_2$ Score (\%)} & \textbf{TS-RGB F$_2$ Score (\%)} \\ \hline
Noise                               & 63.85                         & 45.61                           \\
Ground                              & 70.28                         & ---                             \\
Low Vegetation                      & 51.89                         & ---                             \\
Medium Vegetation                   & 71.82                         & 93.07                           \\
\textcolor{red}{Tower}              & 87.37                         & 63.70                           \\
\textcolor{red}{Power Line}         & 96.05                         & 73.47                           \\\hline
\end{tabular}
\label{tab:fbeta_scores}
\end{table}

The results in Table~\ref{tab:fbeta_scores} show the performance of Point Transformer V3 (PTV3) on the TS40K and TS-RGB datasets. On TS40K, PTV3 achieves excellent F$_2$ scores in key power grid classes, with 87.37\% for towers and 96.05\% for power lines, meeting the performance requirements for reliable inspection tasks. However, on TS-RGB, performance is more variable, with tower segmentation dropping to 63.70\% and power lines to 73.47\%. This variability highlight the challenge of segmenting power grid elements in TS-RGB, likely due to the limited discriminative power of the RGB features for these classes. Nevertheless, PTV3's strong results on TS40K make it a viable tool for power grid inspection.

\section{Conclusion~\label{sec:conclusions}}
This study demonstrates the efficacy of 3D semantic segmentation for automating power grid inspections, addressing key challenges in traditional methodologies. Leveraging the TS40K dataset, we evaluated several state-of-the-art models, achieving substantial improvements in detecting critical grid elements like power lines and supporting towers. Notably, transformer-based models achieved IoU scores exceeding 95\% for power lines, underscoring their potential for deployment in real-world scenarios.
Our findings highlight the transformative impact of machine learning in streamlining inspection workflows, reducing costs, and enhancing grid safety. The integration of uncertainty flagging mechanisms within the proposed inspection tool ensures reliability, balancing automation with human oversight. However, challenges such as handling noisy data and extreme class imbalances remain. Future work will explore hybrid methods that integrate complementary 3D and 2D data modalities, as well as optimizing models for real-time processing in large-scale grid networks.

\section*{Acknowledgments}
\footnotesize
This research was funded by DM 351-2022 of the Italian Ministry of University and Research and with partial support from the Italian INDAM – GNAMPA group, by FCT I.P., through the strategic project NOVA LINCS (UIDB/04516/2020). It was also funded
via the ENFIELD (European Lighthouse to Manifest Trustworthy and Green AI) project, European Union’s Horizon Research and Innovation Programme, Grant Agreement No. 101120657. Views expressed are those of the author(s) only and do not reflect those of the European Union. Neither the European Union nor the granting authority can be held responsible for them.

\ifCLASSOPTIONcaptionsoff
  \newpage
\fi

\bibliographystyle{IEEEtran}
\bibliography{bibtex/bib/IEEEabrv,bibtex/bib/main}








\end{document}